\documentclass[conference]{IEEEtran}
\IEEEoverridecommandlockouts
\usepackage{cite}
\usepackage{amsmath,amssymb,amsfonts}
\usepackage{graphicx}
\usepackage{textcomp}
\usepackage{xcolor}
\usepackage{subfigure}
\usepackage{CJK,balance,url}
\usepackage{soul}
\usepackage{algorithmicx,algorithm,algpseudocode}
\usepackage{enumerate,enumitem,eqnarray}

\def\BibTeX{{\rm B\kern-.05em{\sc i\kern-.025em b}\kern-.08em
    T\kern-.1667em\lower.7ex\hbox{E}\kern-.125emX}}
\begin{document}

\title{
\textit{Keiki}: Towards Realistic Danmaku Generation\\ via Sequential GANs
\thanks{This paper is accepted by the 2021 IEEE Conference on Games and will be included in the proceedings.}
}

\author{\IEEEauthorblockN{Ziqi Wang\IEEEauthorrefmark{1}\IEEEauthorrefmark{2}, Jialin Liu\IEEEauthorrefmark{1}\IEEEauthorrefmark{2}}
\IEEEauthorblockA{\IEEEauthorrefmark{1} \textit{Research Institute of Trustworthy Autonomous System}\\
\textit{Southern University of Science and Technology}}
\IEEEauthorblockA{\IEEEauthorrefmark{2}\textit{Guangdong Provincial Key Laboratory of Brain-inspired Intelligent Computation}\\
\textit{Department of Computer Science and Engineering} \\
\textit{Southern University of Science and Technology}\\
Shenzhen, China\\
11710822@mail.sustech.edu, liujl@sustech.edu.cn}
\and
\IEEEauthorblockN{Georgios N. Yannakakis\IEEEauthorrefmark{3}}
\IEEEauthorblockA{\IEEEauthorrefmark{3}\textit{Institute of Digital Games} \\
\textit{University of Malta}\\
Msida 2080, Malta \\
georgios.yannakakis@um.edu.mt}
}

\maketitle

\begin{abstract}
Search-based procedural content generation methods have recently been introduced for the autonomous creation of bullet hell games. Search-based methods, however, can hardly model patterns of danmakus---the bullet hell shooting entity---explicitly and the resulting levels often look non-realistic. In this paper, we present a novel bullet hell game platform named \textit{Keiki}, which allows the representation of danmakus as a parametric sequence which, in turn, can model the sequential behaviours of danmakus. We employ three types of generative adversarial networks (GANs) and test \textit{Keiki} across three metrics designed to quantify the quality of the generated danmakus. The time-series GAN and periodic spatial GAN show different yet competitive performance in terms of the evaluation metrics adopted, their deviation from human-designed danmakus, and the diversity of generated danmakus. The preliminary experimental studies presented here showcase that potential of time-series GANs for sequential content generation in games. 
\end{abstract}


\section{Introduction}

Bullet hell is a game genre in which a player survives by dodging overwhelming numbers of enemy projectiles and scores by shooting enemies down. The core component of a bullet hell game level is a ``\emph{danmaku}", which refers to the game entity (agent or opponent) that creates bullets and determines their trajectories. Danmakus in bullet hell games are typically controlled by a set of human-designed rules, aiming at the elicitation of challenging and engaging experiences for the player.

\begin{figure}[!tb]
\centering
\subfigure[\label{fig:real}\textit{Touhou Project} series (Team Shanghai Alice, since 2003).]{\includegraphics[width=.8\linewidth, trim=24 0 24 0,clip]{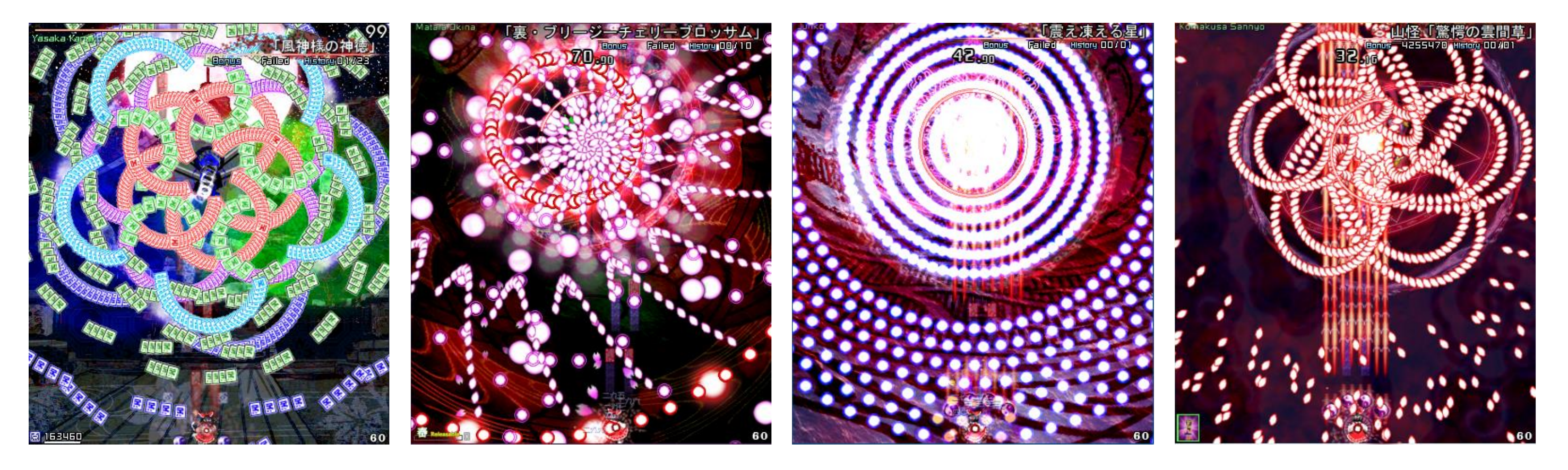}} \\
\subfigure[\label{fig:talakat}\textit{Talakat}: Figure 6 from~\cite{talakat} with authors' permission.]{\includegraphics[width=.8\linewidth, trim=8 0 8 0,clip]{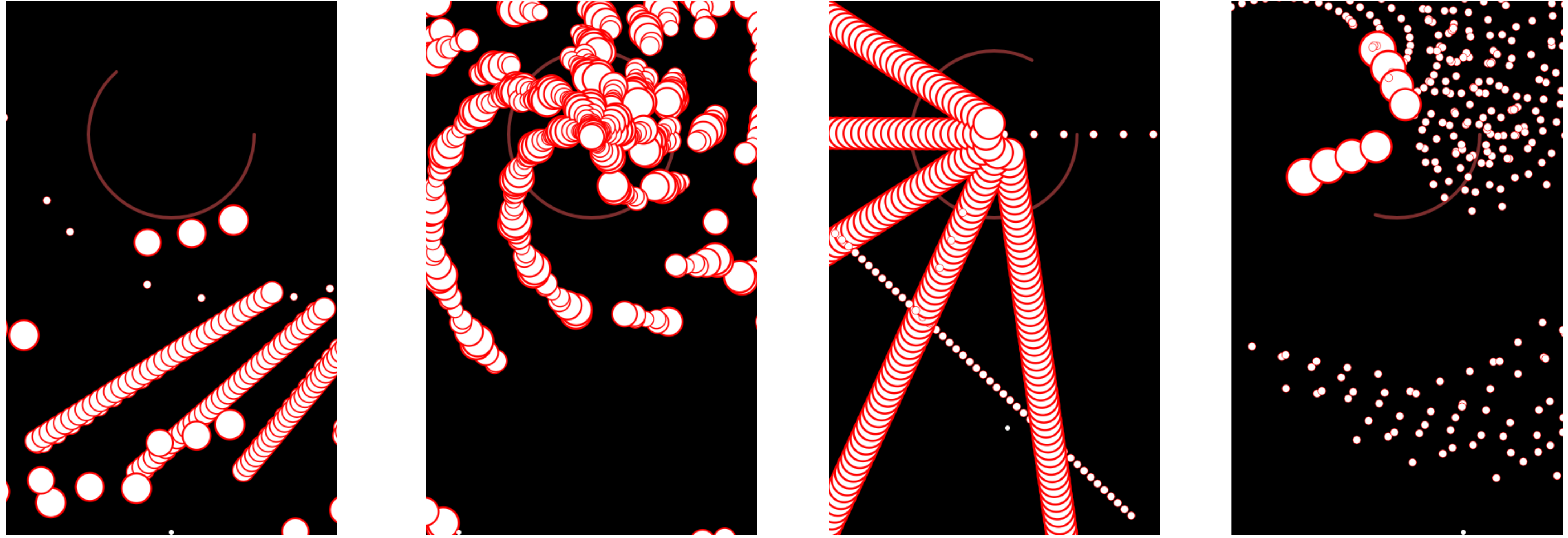}} \\
\subfigure[\label{fig:generatedDCGAN}\emph{Keiki}: danmakus generated by deep convolutional GAN.]{\includegraphics[width=.8\linewidth]{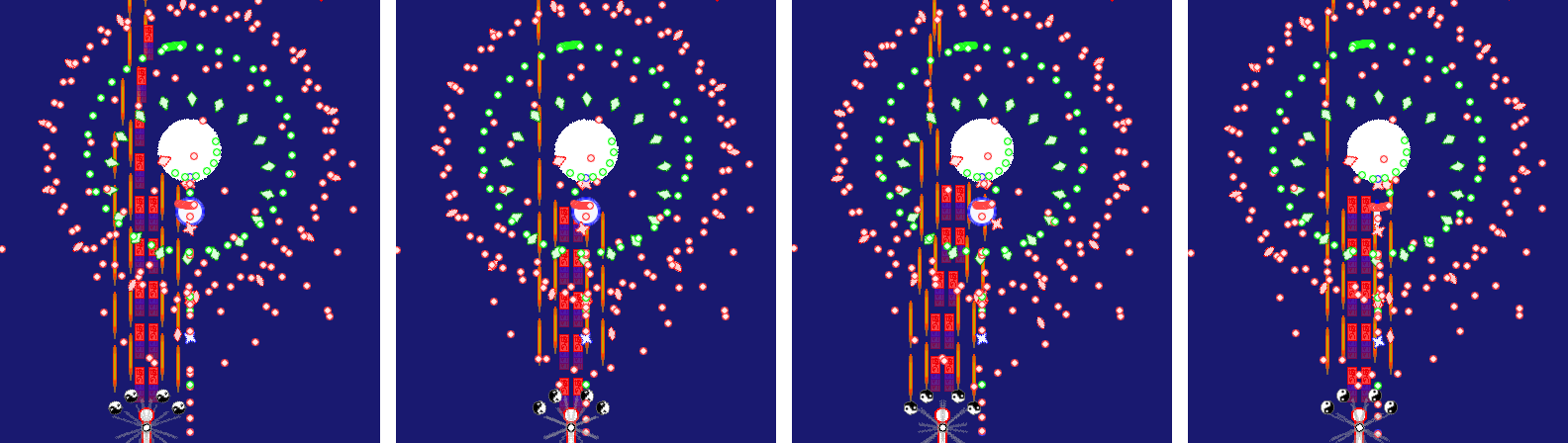}} \\
\subfigure[\label{fig:generatedPSGAN}\emph{Keiki}: danmakus generated by periodic spatial GAN.]{\includegraphics[width=.8\linewidth]{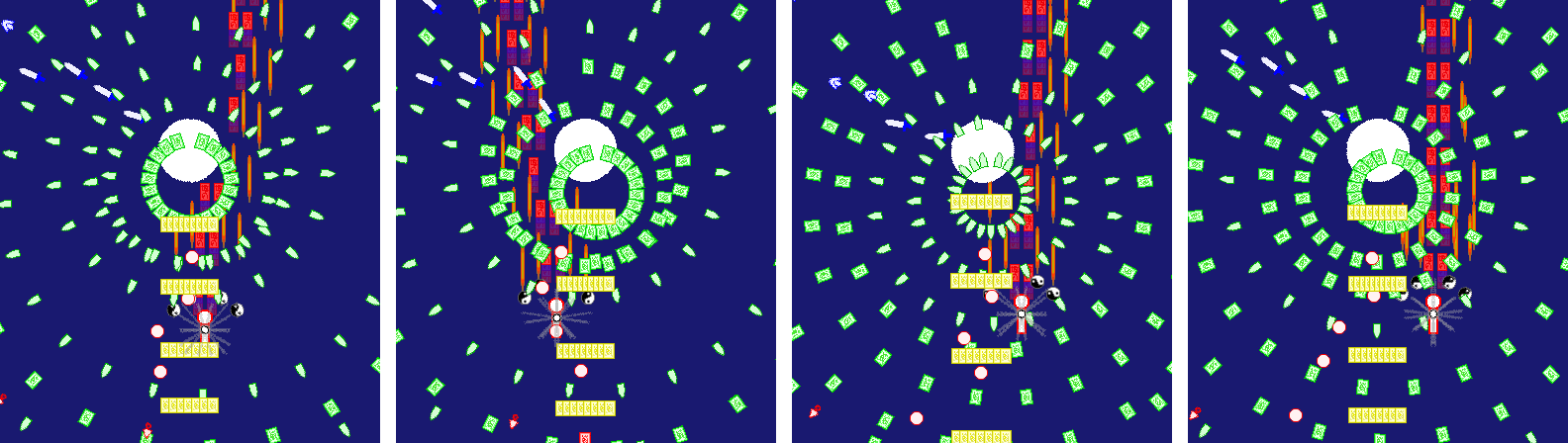}} \\
\subfigure[\label{fig:generatedTimeGAN}\emph{Keiki}: danmakus generated by time-series GAN.]{\includegraphics[width=.8\linewidth]{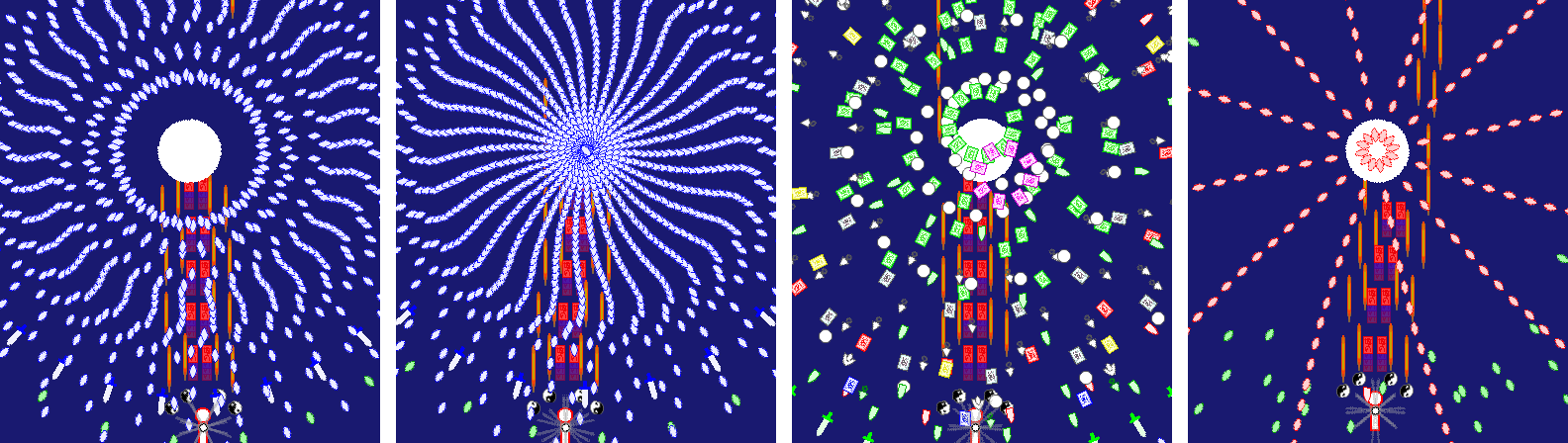} \label{TimeGANgen}}
\caption{Examples of danmakus of commercial games vs. danmakus generated by \textit{Talakat} and by our GAN generators \textit{Keiki} (cf. Section \ref{sec:lg}).}
\label{fig:talakatvsreal}
\end{figure}

Over the last decade, a number of search-based procedural content generation (PCG) approaches have been applied widely to generate levels for platformer games, dungeon-like games and shooting games~\cite{togelius2011search,togelius2013procedural,shaker2016procedural,risi2020increasing,summerville2018procedural,liu2020deep}; however, the generation of bullet hell levels was only been explored recently. \textit{Talakat}~\cite{talakat} first applied PCG to generate bullet hell levels by searching in a human-designed level space represented as a grammar-based text and designed several metrics and constraints to guide the search, although the generated danmakus look rather dissimilar compared to the regular ones of commercial games (cf. Figures \ref{fig:real} and \ref{fig:talakat}). 

Studies linked to bullet hell generation include weapon generation and rhythm game generation. In particular, Hastings \textit{et al.}~\cite{gar} evolved particle-based weapons as compositional pattern producing networks in an online manner and Hoover \textit{et al.}~\cite{hoover2015audioinspace} introduced a similar generative system that considers the background music of the game for weapon particle generation. Constrained surprise search was introduced in~\cite{gravina2016constrained} to generate weapons in first-person shooting games. In \cite{liang2019procedural}, a novel neural network structure combining convolutional neural network (CNN) and recurrent neural network (RNN) was designed to generate rhythm game levels with specific degree of difficulty.

By observing the indicative outcomes of Figures \ref{fig:real} and \ref{fig:talakat}, it appears that search-based PCG methods~\cite{togelius2011search} reach a certain level of quality compared to actual bullet hell levels~\cite{talakat}. Through mere observation, it also seems that existing state-of-the-art approaches can hardly model the implicit danmaku patterns present in commercial-standard games.  In this paper, we present \textit{Keiki}\footnote{The name ``\textit{Keiki}'' comes from a character in \textit{Touhou Kikeijuu $\sim$ Wily Beast and Weakest Creature} (Team Shanghai Alice, 2019).} (cf. Section \ref{sec:keiki}) as a complementary bullet hell game platform that can represent danmakus as parametric sequences and can support neural network-based generative methods. \textit{Keiki} currently employs three types of generative adversarial networks (GANs)~\cite{gan}. The various GAN models are evaluated across three metrics designed to quantify the quality of the generated danmakus.

\section{\textit{Keiki}: Bullet Hell Platform Generator}\label{sec:keiki}

The \textit{Keiki} platform provides functions including danmaku design, encoding, evaluation and basic gameplay. The source code of \textit{Keiki} is available on Github\footnote{\url{https://github.com/PneuC/Keiki}}, including the training data and all image assets used in this work. Due to the page limit, we only describe the danmaku encoding as follows.

The design of a danmaku can be seen as specifying a tuple $\langle F, \mathbf p \rangle$, where $F$ refers to the shooting rules and the vector $\mathbf p$ specifies the values of $F$'s control parameters, usually defined in the construction method of the implementation class. At the $t^{th}$ frame, the danmaku calls the implemented bullet builder several times according to its shooting rules ($F$) and parameter values ($\mathbf p$). This process is denoted by $F(t | \mathbf p)\mapsto (\mathbf x_1, \mathbf x_2, \cdots, \mathbf x_{
N^t})$, where $N^t\in \mathbb{N}$ is the number of times that this danmaku calls the bullet builder at the $t^{th}$ frame and $\mathbf x_i$ is the parameter vector used at the $i^{th}$ call. When shooting multiple times is desired (e.g., $L$ times), $F(t | \mathbf p)$ will be repeatedly called for $t \in \{1, 2, \cdots\}$ until $t = T$ such that $\sum_{t=0}^{T} N^t \geq L$ is satisfied. Thus, the length of the parametric sequence to be generated is $L$.
To reduce the dimensionality and data redundancy, the data compress method proposed in the work of \cite{crnngan} is used, described as follows. A parameter $itv$ (short for interval, corresponding to $time$ in \cite{crnngan}), representing the time passed since the last call of bullet builder, is also added to the parametric sequence.

\section{Danmaku Generation}\label{sec:lg}

In Section \ref{sec:lggan} we outline the three GAN methods used for danmaku generation. Section \ref{sec:metrics} introduces the metrics we used to evaluate the danmaku performance. Our preliminary experimental results are presented and discussed in Section \ref{sec:xps}.

\subsection{Danmaku Generation with GANs}\label{sec:lggan}

Deep convolutional GANs, periodic spatial GANs and time-series GANs are considered separately and employed to generate parametric sequence of length 64. Generators are trained on a dataset of 34 danmakus implemented by the first author of this paper. These danmakus are mainly imitations of danmakus found in \textit{Touhou Project}\footnote{\url{https://touhou.fandom.com/wiki/Touhou_Project}} series. 

During training, data augmentation is applied as follows. Each time a danmaku $\langle F, \mathbf p \rangle$ is loaded from training data, a Gaussian mutation is added to its parameters, i.e., $\mathbf p \gets \mathbf p + 0.05 ~ \mathcal N (0,\mathbf p \times \mathbf I )$, before feeding it into the model.

\subsubsection{Time-series GAN}

We implement time-series GAN (TimeGAN)~\cite{TimeGAN} as an alternative PCGML approach, as it combines autoencoder and supervised loss to help learning temporal dynamics from training data. 
Not only a generator and a discriminator are involved in TimeGAN, but also a pair of an embedder (encoder) and a reconstructor (decoder), trained with a reconstruction loss, a supervised loss and an adversarial loss jointly. For all these models in TimeGAN, a 3-layer stacked LSTM with 128 hidden units in each layer is used and the logistic function is employed in the output layer. 
The dimensionality of the autoencoder's hidden space is set as 24. 
The noise input to TimeGAN is composed of a global noise $Z^g$ and a periodic noise $Z^p$. For any $i \in \{1,2,\dots, T_Z\}$, the input noise is $Z^g \oplus Z^g_i$, where $\oplus$ denotes concat. $Z^g$ is sampled once and duplicated $T_Z$ times ($T_Z$ is the spatial length of the noise, set as $10$ here), while $Z^p$ is sampled as $Z^p_i = sin(i \cdot K_1(Z^g) + K_2(Z^g))$, where $K_1$ and $K_2$ are two multi-layer perceptrons.

\subsubsection{Periodic spatial GAN}

The periodic spatial GAN~\cite{psgan} with sequential noise input is employed as our second, alternative, generative model. The noise input to periodic spatial GAN is the same as that of TimeGAN.
We use four one-dimensional convolutions with no padding. The kernel size and stride are set as $(4, 1), (4, 2), (4, 1), (4, 2)$, respectively, for each layer. Such a design guarantees that the generator will create sequences of length $64$ when the spatial length of input noise is $10$, which is directly comparable with other employed GAN architectures. The structure of the discriminator is symmetric to the generator. Both the generator and the discriminator use the logistic activation function in the final layer and the ReLU activation function in all other layers.

\subsubsection{Deep convolutional GAN}

A vanilla deep convolutional GAN (DCGAN)~\cite{dcgan} is implemented as a non-sequential baseline. We used 5 one-dimensional transposed convolutional layers in the generator. The kernel size, stride and padding of the first and the remainder layers are $(4, 1, 0)$ and $(4, 2, 1)$, respectively. The number of output channels of the first layer is $256$ and are reduced by $2$ times at each layer that follows. The structure of the discriminator is also symmetric to the generator, and the activation function is the same as the one used in the periodic spatial GAN.

\subsection{Evaluation Metrics}\label{sec:metrics}

Three feature-based metrics are designed to evaluate a generated danmakus $D$. The first one is \emph{shooting frequency}, $SF$, which determines whether all bullets are shot in a very short period or progressively.
    \begin{equation}
        \label{eq:sf}
        SF(D) = L / T,    
    \end{equation}
\noindent where $L$ is the number of shooting times (i.e., the length of parametric sequence) and $T$ is danmaku's duration. 

The second metric used is \emph{mean momentum}, defined as the sum of all bullets on the screen over time. Let $B^t$ be the set of all the bullets at the $t^{th}$ frame. For any bullet $b$, $b_w$ and $b_s$ denote respectively its weight (a scalar value proportional to the size of its image) and its speed. The mean momentum of danmaku $D$ is calculated as:
    \begin{equation}
        \label{eq:mm}
        MM(D) = \frac{1}{T} \sum\limits_{i = 1}^{T} \sum\limits_{b \in B^t} b_w b_s,
    \end{equation}

\noindent where $T$ is the danmaku's duration.
    
\emph{Coverage} is another metric used to estimate the degree of game difficulty. The game screen is split into grids of $8*8$ pixels and the percentage of regions that is covered by any bullet at any frame is computed. Let $r$ and $c$ be the number of rows and columns of the regions, respectively. $cov^t_{i, j}$ determines whether the region at position $i,j$ is covered by at least one bullet at $t^{th}$ frame. The coverage metrics is as follows:.
    \begin{equation}
        \label{eq:c}
        C(D) = \frac{1}{rc} \sum\limits_{i = 1}^{r} \sum\limits_{j = 1}^{c} \mathbf{1}(\exists~ t \in [1, T], cov^t_{i, j}),
    \end{equation}
\noindent where $\mathbf{1}(p)$ is $1$ if the proposition $p$ holds, otherwise $0$. Higher $C$ values imply more difficult levels.

Although the aforementioned metrics may not suffice to determine if generated danmakus are of good quality, they can be used to quantify the similarity between real and generated danmakus with distance measures, such as those of the KL-divergence family.

\subsection{Experimental Study and Discussion}\label{sec:xps}

After preliminary hyper-parameter tuning, the following setting is used in our experiments.
The batch size is set to $12$ for all GANs. The learning rate is $2*10^{-4}$ for the DCGAN and the periodic spatial GAN, while it is set to $2*10 ^{-3}$ for the TimeGAN. We optimise the DCGAN and the periodic spatial GAN via Adam and the TimeGAN via RMSprop.
The DCGAN and periodic spatial GAN are trained separately for $5,000$ iterations. The autoencoder of TimeGAN is pre-trained for $5,000$ iterations, then the generator is trained with supervised loss only for $500$ iterations; finally we the trained all the models jointly for $5,000$ iterations.

\begin{figure}[htbp]
    \centering
    \includegraphics[width=\linewidth]{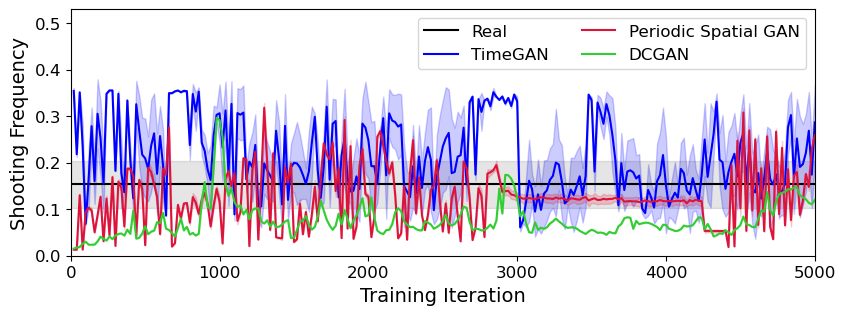}\hfill
    \includegraphics[width=\linewidth]{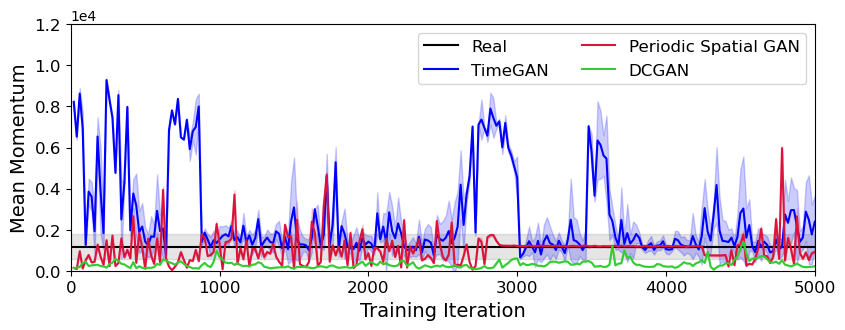}\hfill
    \includegraphics[width=\linewidth]{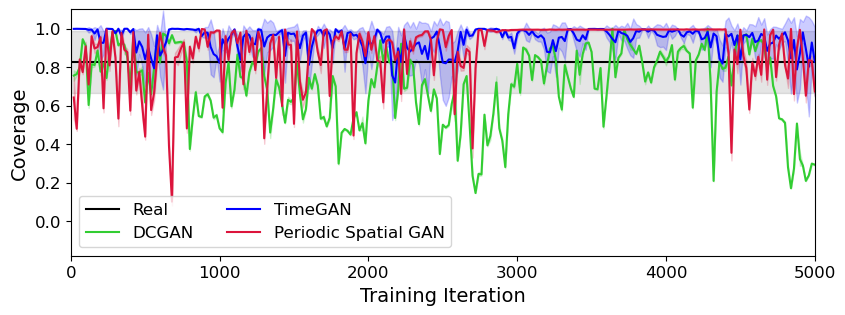}
    \caption{Comparison of GANs on shooting frequency (top), mean momentum (middle) and coverage (bottom) during training. Every 20 iterations, the trained GAN generates 30 samples for evaluation. Solid curves depict the mean value over 30 samples and shadow areas illustrate the standard derivation.}
    \label{fig:curves}
\end{figure}

Figure \ref{fig:curves} shows how the values of metrics introduced in Section \ref{sec:metrics} change during training. The DCGAN performs the worst since its metric values are the ones furthest to the ones from real data. The periodic spatial GAN, on the other hand, shows more stable performance compared to that of the TimeGAN. The instability of TimeGAN in our experiments may be explained by its complex architecture which, in turn, calls for larger amounts of training data. According to Figure \ref{fig:curves}, however, the TimeGAN approach has the highest standard derivation across all metrics, which implies the highest diversity. The other GANs appear to suffer from mode collapse, i.e., they only generate data of one or a limited number of patterns. 

Figures \ref{fig:generatedDCGAN}, \ref{fig:generatedPSGAN} and \ref{fig:generatedTimeGAN} show indicative examples of danmakus generated by the different GANs.
DCGAN generates almost similar danmakus; periodic spatial GAN generates danmakus of similar patterns whereas TimeGAN appears to generate far more diverse danmakus compared to the other two GANs. 

\section{Further Discussion}

Generating danmakus via PCGML \cite{summerville2018procedural} using GANs is a challenging task; especially when training data is not readily available. Bootstrap methods~\cite{bootstrappcg} or PCG via reinforcement learning (PCGRL)~\cite{khalifa2020pcgrl,shu2021edrl} are worth investigating to overcome this challenge, but collecting more human-designed danmakus as training data is also important. While implementing and coding danmakus is time-consuming, videos of danmakus are easier to obtain from real games; thus, learning to represent danmakus directly from videos appears to be a promising future direction. The design of alternative GAN architectures is also important to be investigated for future work. 

Another important direction for future research is agent-based evaluation.
An $A^*$ agent has already been implemented in \textit{Keiki}. Agent-based testing, however, is currently inefficient due to the slow simulation times. In bullet hell games hundreds of bullets may exist simultaneously at each frame; in \textit{Keiki}, bullets' moving directions can vary depending on player's current or previous positions, which makes the real-time simulation CPU intensive. If more efficient agents are implemented or the game simulation can eventually be accelerated, we plan to add constraints regarding the playability of generated danmakus by using techniques such as constrained adversarial nets~\cite{cans}. 

The three metrics currently used this work are intuitively designed. Designing new evaluation metrics and employing other PCG methods, such as that in \cite{talakat}, as baselines of the \emph{Keiki} platform define top priorities for future work.

\section{Conclusion}

In this paper, we presented \textit{Keiki}, a novel bullet hell platform which allows encoding danmakus into a parametric sequence so that various artificial level generators can be used to generate danmakus. We also introduce GANs~\cite{gan} as an alternative PCGML \cite{summerville2018procedural} mechanism that learns to represent human-designed danmakus---and hence generates more realistic content (cf. Section \ref{sec:lg})---and employ three different GAN architectures to generate them.
Experimental results on \textit{Keiki} show that both the TimeGAN and the periodic spatial GAN methods have the potential of generating realistic danmakus. The TimeGAN, in particular, generated more diverse danmakus, while the periodic spatial GAN is not sensitive to the length of parametric sequence since it is based on CNNs. Danmaku generation with GANs can be further enhanced though agent-based evaluation and constrained optimisation to ensure playability and yield more realistic danmakus that feature particular human-designed patterns or features. 

\section*{Acknowledgement}
The authors thank the reviewers for their careful reviews and insightful comments.

\balance

\bibliographystyle{IEEEtran}
\bibliography{keiki,dlpcg}

\end{document}